\documentclass[10pt,twocolumn,letterpaper]{article}

\usepackage[pagenumbers]{wacv} 

\usepackage{graphicx}
\usepackage{amsmath}
\usepackage{amssymb}
\usepackage{booktabs}
\usepackage{times}
\usepackage{epsfig}
\usepackage{graphicx}
\usepackage{amsmath}
\usepackage{amssymb}
\usepackage{pgfplots}
\pgfplotsset{width=7cm,compat=1.16}

\usepackage{tikz}
\usepackage{graphicx}
\usetikzlibrary{positioning}
\usepackage{textcomp, gensymb}

\newcommand{\FF}{\mathcal{F}}
\newcommand{\TT}{\mathcal{T}}
\newcommand{\PP}{\mathcal{P}}
\newcommand{\DD}{\mathcal{D}}

\usepackage{pifont}
\usepackage[pagebackref,breaklinks,colorlinks]{hyperref}

\usepackage[capitalize]{cleveref}
\crefname{section}{Sec.}{Secs.}
\Crefname{section}{Section}{Sections}
\Crefname{table}{Table}{Tables}
\crefname{table}{Tab.}{Tabs.}

\def\wacvPaperID{2672} 
\def\confName{WACV}
\def\confYear{2024}

\begin{document}

\title{Towards Realistic Out-of-Distribution Detection: A Novel Evaluation Framework for Improving Generalization in OOD Detection}


\author{
  Vahid Reza Khazaie\thanks{Vector Institute, Toronto, Canada}
  \and
  Anthony Wong\thanks{Western University, London, Canada}
  \and
  Mohammad Sabokrou\thanks{Okinawa Institute of Science and Technology, Onna, Japan}
}

\maketitle

\begin{abstract}
This paper presents a novel evaluation framework for Out-of-Distribution (OOD) detection that aims to assess the performance of machine learning models in more realistic settings. We observed that the real-world requirements for testing OOD detection methods are not satisfied by the current testing protocols. They usually encourage methods to have a strong bias towards a low level of diversity in normal data. To address this limitation, we propose new OOD test datasets (CIFAR-10-R, CIFAR-100-R, and ImageNet-30-R) that can allow researchers to benchmark OOD detection performance under realistic distribution shifts. Additionally, we introduce a Generalizability Score (GS) to measure the generalization ability of a model during OOD detection. Our experiments demonstrate that improving the performance on existing benchmark datasets does not necessarily improve the usability of OOD detection models in real-world scenarios. While leveraging deep pre-trained features has been identified as a promising avenue for OOD detection research, our experiments show that state-of-the-art pre-trained models tested on our proposed datasets suffer a significant drop in performance. To address this issue, we propose a post-processing stage for adapting pre-trained features under these distribution shifts before calculating the OOD scores, which significantly enhances the performance of state-of-the-art pre-trained models on our benchmarks.
\end{abstract}

\section{Introduction}
\label{sec:intro}

\begin{figure}[tb]
\begin{center}
   \includegraphics[width=0.42\textwidth]{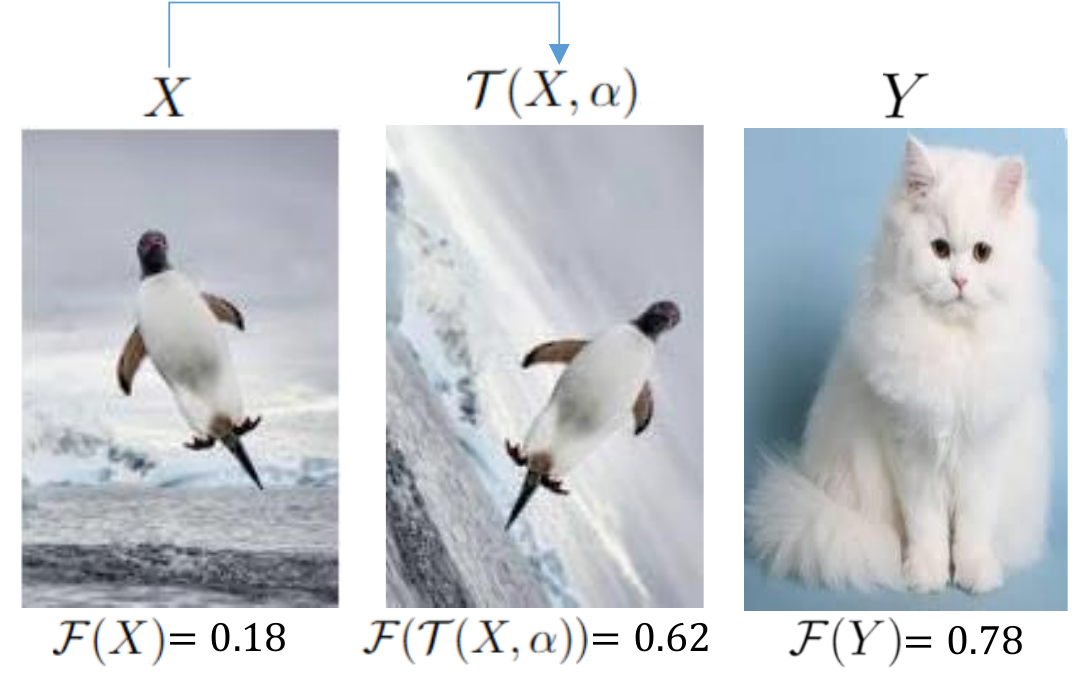}
\end{center}
\vspace{-4mm}
   \caption{Example of an out-of-distribution (OOD) detector $\FF_{\theta}$ trained to detect Penguins.  $\FF_{\theta}$ is a state-of-the-art OOD detector that recognizes $X$ as an inlier and $Y$ as an OOD sample with high confidence but fails to recognize $\TT(X,\alpha)$ as an inlier which is modified by a semantic-preserving transformation. As shown in the figure,  $\FF_{\theta}(X) << \FF_{\theta}(\TT(X,\alpha))$ and $\TT(X,\alpha)$ will labeled as an OOD sample. This situation raises concerns about the reliability and safety of OOD detection methods, as it compromises the ability of the detector to accurately distinguish between inliers and OOD samples.}
\label{fig:fig1}
\end{figure}

Detecting samples that deviate from the norm is the primary goal of out-of-distribution (OOD) detection. When the distribution of normal data contains multiple semantic concepts, identifying OOD samples by learning intrinsic concepts of in-distribution (ID) samples becomes an extremely challenging task. The inability to identify such samples accurately poses significant challenges to the reliability and trustworthiness of machine learning methods. As a result, OOD detection is a crucial task for the development of reliable and trustworthy machine learning systems\cite{salehi2021unified,yang2021generalized}.

This paper focuses on leveraging deep pre-trained features for OOD detection as they are semantically richer, having been extracted from models trained on large and diverse datasets. Unlike other OOD detection methods that rely on training on inlier data, such deep pre-trained features can capture multiple semantic concepts. Deep models trained on large-scale datasets can be repurposed for different tasks with minimal fine-tuning. We hypothesize that with some adaptation, deep pre-trained features extracted from these models can perform better on our proposed evaluation framework.

In recent years, a vast amount of research has been conducted that has focused on improving only the performance on OOD detection. Methods such as DN2 \cite{Bergman2020DeepNN}, CSI \cite{Tack2020CSIND}, ODIN \cite{Liang2018EnhancingTR}, 
FITYMI \cite{mirzaei2022fake} and PANDA \cite{Reiss2021PANDAAP} have saturated performance on standard existing OOD benchmarks, indicating that this field has reached its peak. However, this raises the question of whether current state-of-the-art (SOTA) methods are effective in real-world settings. \textit{\textbf{We believe that it is time to take a step back and analyze the status quo of this research area}}.

We observe that the requirements of real-world OOD detection methods are not reflected by current testing protocols. Current SOTA methods have this inductive bias that normal samples have very similar distribution to the training set during test time while anomalies are distributed much further. However, real-world test samples often contain various levels of distribution shift while maintaining semantic consistency. As an example, we can think of a factory that expects that a model produces the same prediction for a normal screw and a screw transformed by a semantic-preserving geometric function like ~$\TT$. However, in practice, we will see that $\TT$ can cause distribution shifts and compromise the performance of SOTA OOD detectors (see Fig. \ref{fig:fig1}).

This unreliability of OOD methods is due to the above-mentioned inductive bias.  In fact, the current SOTA is vulnerable to $\TT$. Looking at current testing protocols, such distribution shifts are not taken into account in current benchmark datasets like CIFAR-10 and CIFAR-100 \cite{krizhevsky2009learning} which have low diversity of in-distribution (ID) samples and semantically distant OOD samples. Thus, the flawed testing protocol encourages methods to have a strong bias towards a low level of diversity in normal data, which is detrimental to the methods' real-world deployability. For example, in some SOTA methods, such as CSI\cite{Tack2020CSIND} the decision boundary lies extremely close to the ID samples, allowing the method to easily detect the far OOD samples. This is a consequence of using transformations of inliers as OOD samples with the contrastive learning paradigm. As a result, they ignore the variations within the inlier sets. Consequently, under realistic conditions, we believe that these assumptions are insufficient to develop reliable methods. Moreover, future research that simply improves performance on existing OOD benchmarks is not advancing the field toward more real-world applicable OOD detection methods. \textit{\textbf{Therefore, this paper introduces a framework that addresses generalization within the context of OOD detection, thereby bridging a gap between these two fields.}}
%

%


\begin{figure}[t]
\begin{center}
   \includegraphics[width=0.5\textwidth]{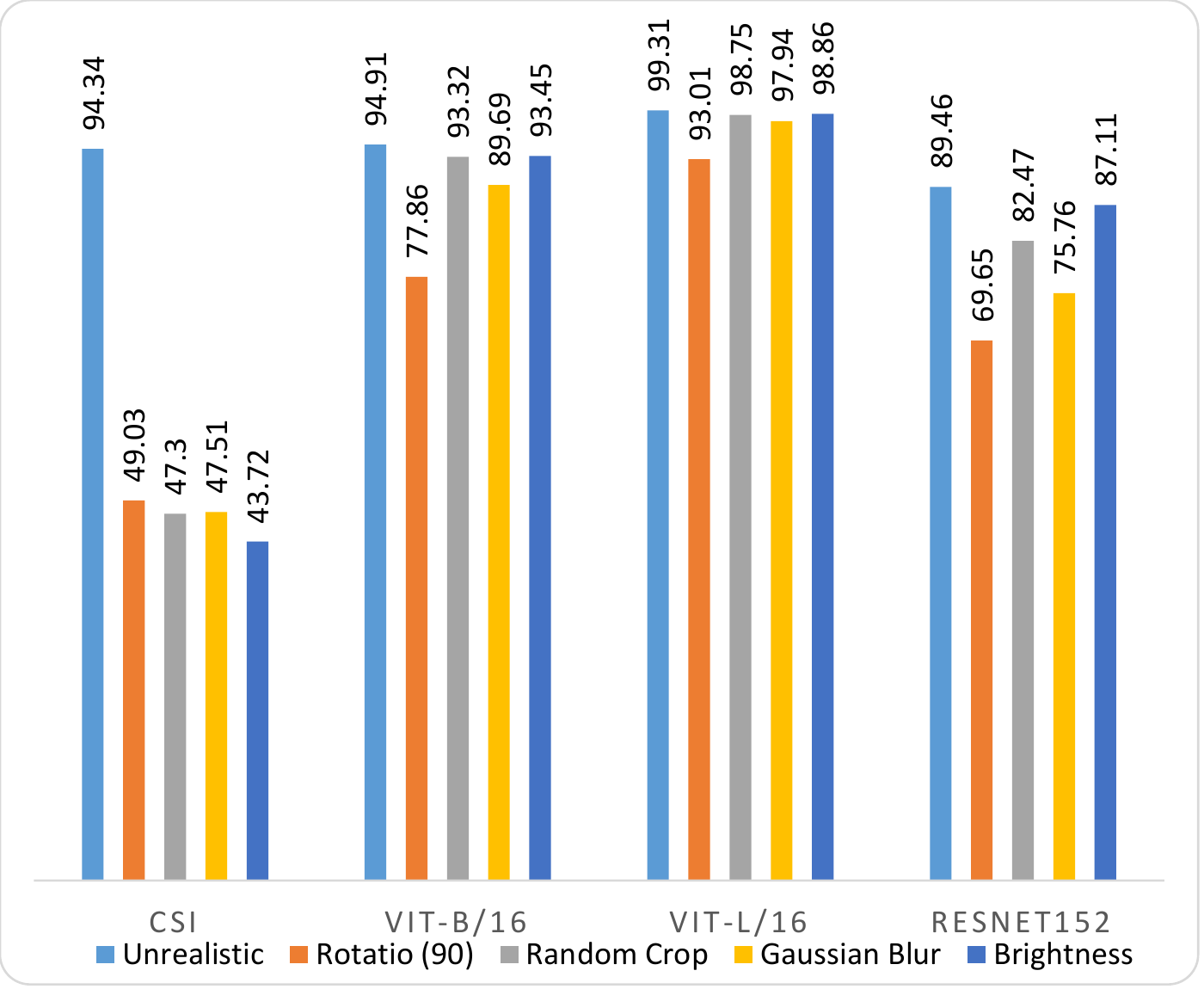}
\end{center}
\vspace{-4mm}
   \caption{This figure compares the performance of multiple SOTA models. The light blue bars display the performance of these models on existing OOD detection benchmarks, whereas the other bars display the performance of the same method on the same dataset but also include samples transformed by $\TT$. We can observe a drastic drop in performance caused by a single $\TT$. This clearly displays the consequence of inductive bias in SOTA methods as a result of existing testing protocols. Therefore, metrics reported by existing evaluation frameworks do not represent real-world performance.}
\label{fig:fig2}
\end{figure}



    



At its core, changing the direction of research requires a testing protocol that represents more realistic conditions and an evaluation metric to quantify how well the problem has been solved. To better reflect the conditions in the real world, we propose a new benchmark evaluation framework consisting of new OOD detection benchmark test sets and a new metric. To create our test sets, one potential method of increasing diversity in distribution shifts while maintaining semantic meaning is to combine existing datasets. However, this is limited by the fact that finding semantically matching classes across datasets is not always possible for research and real-world applications. Thus, we must create an OOD detection benchmark that simultaneously contains sufficient intra-class distribution shift while maintaining semantic meaning. This can be achieved using common corruptions \cite{Hendrycks2019BenchmarkingNN} and data augmentations. Common corruptions are purposefully designed image perturbations that mimic frequently encountered perturbations in natural images\cite{Hendrycks2019BenchmarkingNN}. Models are expected to be robust to images with common corruptions because such samples maintain semantic meaning. Augmentations are realistic transformations applied to the data to increase its diversity without disturbing the semantic meaning of the transformed image. Our benchmark is created from existing datasets, such as CIFAR-10 \cite{krizhevsky2009learning}, CIFAR-100 \cite{krizhevsky2009learning}, and ImageNet-30 \cite{hendrycks2019selfsupervised}, by applying semantic-preserving transformations. In summary, augmentations and common corruptions can be applied to current datasets to produce a new OOD detection benchmark that more accurately reflects the suitability of methods for real-world scenarios. Our evaluation metric which is called Generalizability Score (GS) measures the difference in performance between previous benchmarks and our proposed benchmark. If a method can generalize to all in-class distribution shifts, the difference will be zero. The proposed evaluation framework simultaneously measures both the ability of the methods to detect OOD samples and their ability to generalize to semantically-transformed ID samples. As a result, we suggest that future OOD detection methods follow this research avenue.

\textit{\textbf{The main focus of this paper is to establish a framework to simultaneously assess the performance and reliability of OOD detection methods when data have undergone semantic-preserving transformations. The primary objective is to prevent the semantically-consistent transformed samples from being erroneously identified as OOD samples.}}




\section{Related Works}
\label{sec:related_works}
Reconstruction-based anomaly detection is a classical approach that uses the training set to learn patterns that reflect the normal data in an effective way. Based on the learned semantic features, they attempt to reconstruct a new sample at test time. The method assumes that normal data will be reconstructed well, while abnormal data cannot. Samples are classified as normal or anomalous based on thresholds applied to reconstruction error. For instance, a model for video outlier detection suggested by Cong et al. \cite{cong2011sparse} included sparse representations to distinguish inliers from outliers. Using representations learned from inlier data and the reconstruction error, \cite{xu2015learning, sabokrou2016video} detects out-of-distribution data. Several deep learning models with encoder-decoder architecture have also used this score to detect anomalies \cite{sakurada2014anomaly, zhai2016deep, zhou2017anomaly, zong2018deep, chong2017abnormal}. While these methods are effective, they are limited by their poorly designed latent space. 

In addition, adversarial training can be utilized to detect out-of-distribution data. By combining Generative Adversarial Networks (GANs) \cite{goodfellow2014generative} with denoising autoencoders, Sabokrou et al. \cite{sabokrou2018adversarially} proposed a one-class classifier for novelty detection which uses the discriminator's score for reconstructed samples. By modifying the discriminator role to distinguish between good and bad reconstruction quality, Zaheer et al. redefined the adversarial one-class classifier training setup to improve its results \cite{zaheer2020old}. In \cite{Zaheer_2022_CVPR}, a new unsupervised generative learning approach for video anomaly detection exploits the low frequency of anomalies by building cross-supervision between generators and discriminators. To force normal samples to be distributed uniformly across the latent space, Perera et al. used denoising auto-encoder networks \cite{perera2019ocgan} in an adversarial manner. In \cite{jewell2022one}, an adversarial setup is utilized to mask the input of the autoencoder intelligently and learn more robust representations. A deep autoencoder with a parametric density estimator is proposed by \cite{abati2019latent} with an autoregressive procedure to learn its latent representations. In spite of showing success in some OOD detection scenarios, training instability is a limitation of this category of methods.

There are also methods that use contrastive and self-supervised techniques. For self-supervised anomaly detection, Rot uses an auxiliary task of rotation prediction. By using rotation-prediction methods, GOAD \cite{bergman2020classification} learns a feature space in which the inter-class separation between normal data is relatively small. Methods such as CSI \cite{Tack2020CSIND} leverages contrastive learning paradigm to contrast against distribution-shifted augmentations of the data samples along with other samples. In SSD \cite{sehwag2021ssd}, anomalies are scored by the Mahalanobis distance based on K-means clusters. These types of methods may perform well on current benchmarks, but they have some unrealistic assumptions about out-of-distribution data when augmentations such as rotation are applied. This can compromise their reliability in cases where the semantic meaning remains consistent over distribution shifts of inlier data.

A very promising direction in OOD detection is to leverage deep pre-trained features. Pretrained features, when combined with simple anomaly detection methods, achieve superior performance compared to complex state-of-the-art methods. A method that works based on pre-trained features is DN2 \cite{Bergman2020DeepNN} which estimates density using deep pre-trained features and nearest neighbor. Each sample is scored according to the distance from its nearest normal training image. In the case of a larger distance, there is a lower density of normal samples, so an abnormality is more likely to occur. PANDA \cite{Reiss2021PANDAAP} is an anomaly detection method that achieves state-of-the-art performance by leveraging deep pre-trained features and proposing techniques to combat feature deterioration and adapt the deep pre-trained features to the target distribution. In this paper, we assess the performance of deep pre-trained features on our proposed testing framework using a distance-based approach. Furthermore, we propose to adapt these features for our testing framework. Unlike PANDA, our method does not necessitate joint optimization or training and is computationally more efficient.

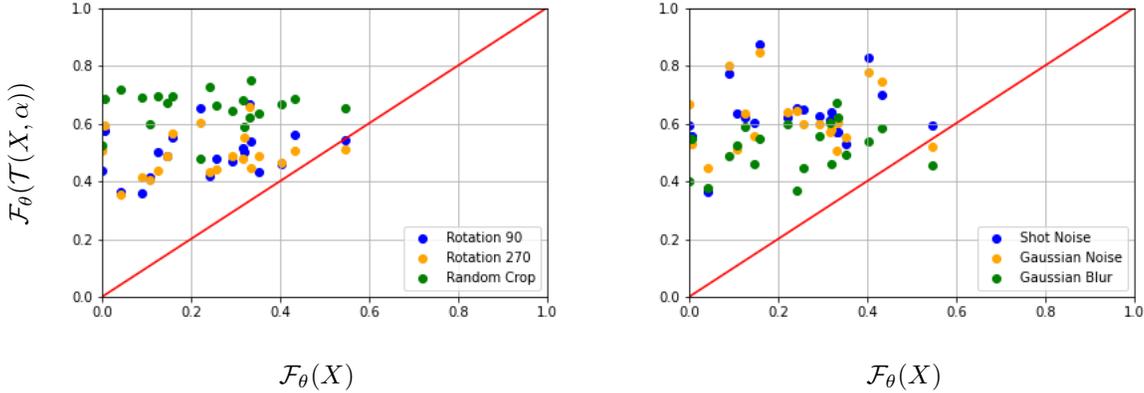
\begin{figure*}[htb]
\begin{minipage}{0.4\textwidth}
\begin{tikzpicture}
  \node (img)  {\includegraphics[scale=0.5]{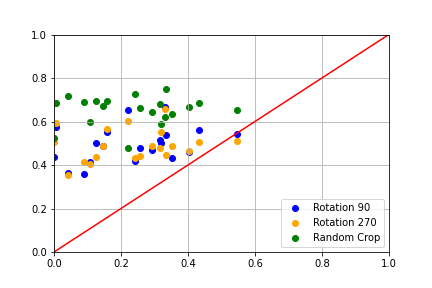}};
  \node[below=of img, node distance=0cm, yshift=1cm,font=\color{black}] {$\FF_{\theta}(X)$};
  \node[left=of img, node distance=0cm, rotate=90, anchor=center,yshift=-1cm,font=\color{black}] {$\FF_{\theta}(\TT(X,\alpha))$};
 \end{tikzpicture}
\end{minipage}%
\begin{minipage}{0.4\textwidth}
\begin{tikzpicture}
  \node (img)  {\includegraphics[scale=0.5]{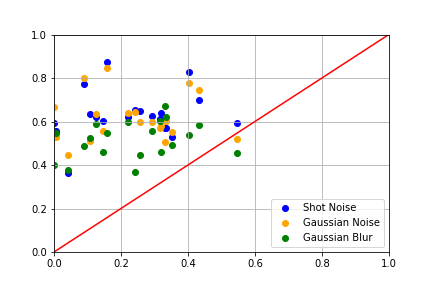}};
  \node[below=of img, node distance=0cm, yshift=1cm,font=\color{black}] {$\FF_{\theta}(X)$};
  \node[left=of img, node distance=0cm, rotate=90, anchor=center,yshift=0cm,font=\color{black}] {};
\end{tikzpicture}
\end{minipage}%
\caption{The figures plot the OOD scores of inlier samples from the CIFAR-10 test set against the OOD scores of the same samples transformed by augmentations or common corruptions, showing that the OOD score for most samples increases towards one after transformation. This indicates that current SOTA models have low generalizability on such transformations, which do not change the semantic meaning of the data. An ideal and robust model should produce the same score for an inlier sample even after transformation.}
\label{fig:fig3}
\end{figure*}



    




\section {Realistic OOD Detection}
\label{sec:method}
In this section, we provide a detailed description of a new evaluation framework that can be applied to more realistic scenarios. To achieve this, we introduce new datasets for OOD detection. Additionally, we demonstrate that SOTA OOD detection methods fail on these datasets. Finally, as a more reliable solution,  we propose to reduce the sensitivity of deep pre-trained features to non-semantic changes in input (e.g., rotation) through adaptation.

\textbf{Proposed Evaluation Framework}
To begin, we define the existing OOD detection benchmark setup. Inlier training data is represented as $Q=[Q_{1}...Q_{n}]$. Combining the inlier test data $B=[B_{1}...B_{n}]$ and OOD test data $Y=[Y_{1}...Y_{n}]$, the test set $S$ is formulated by their union: $S = \{ Y \cup B \}$ where  $B \approx \PP_{I}$ and $Y \not\approx \PP_{I}$. We define $\PP_{I}$ as the distribution of inliers. In general, training the OOD detection method involves optimizing the parameters $\theta$ of a function such as $\FF_{\theta}(X)$ such that it outputs an OOD score of a sample $X$. Similarly, conventional OOD methods are optimized on $Q$ and detect the OOD samples by thresholding on the output of learned $\FF_{\theta}$. As previously mentioned, current methods are designed based on hypothesis $\mathbf{H}$ which is defined in Equ.~\ref{eqH}.   

\begin{equation}
\label{eqH}
    \begin{split}
     \mathbf{H}: \forall Q_i, B_i\approx\PP_{I}  \text{ and }  \forall Y_i\not\approx\PP_{I}  \\
      \mathcal{D}(Q_i, B_i) << \mathcal{D}(Q_i, Y_i)
    \end{split}
\end{equation}
 
Where,  $\DD$ computes the whole difference (i.e, consider both semantic and non-semantic features) between two images. In a real-world application, it is expected that  $\DD$ utilizes only semantic features to make decisions. Being sensitive to the non-semantic features leads to failure when the model faces ID samples altered using semantic preserving transformations.

Here, we propose a new benchmark that better represents real-world conditions and demonstrates that current SOTA methods fail in such a benchmark. An ideal and robust OOD detector should output the same OOD score i.e $\FF_\theta(X)$ for an inlier sample that is non-semantically changed by some transformations $\alpha$ (e.g, $\alpha$ can be a rotation). Ideally, the OOD score must be the same i.e., (($\FF_{\theta}(X) - \FF_{\theta}(\TT(X, \alpha)) < \epsilon)$) where $\epsilon$ is a very small value close to zero. In other words, the method should be robust to transformations such as $\TT$.   In this paper, we show that most of the deep pre-trained models fail in the above-mentioned setup. This means (($\FF_{\theta}(X) - \FF_{\theta}(\TT(X, \alpha)) > > \epsilon$). If the difference between $\FF_{\theta}(X)$ and $\FF_{\theta}(\TT(X,\alpha))$ is large,  $\TT(X,\alpha)$ will very likely be mistakenly labeled as OOD, which compromises the reliability of OOD detection methods in the presence of $\alpha$. Therefore,  we introduce new realistic benchmarks and define a new score for evaluating the effectiveness of models to consider both detection performance and generalizability.
To compare the performance of SOTA methods between our benchmarks versus current benchmarks, we define a geometric transformation function $\TT(B, \alpha)$ parameterized by a sample $X$ and transformation set $\alpha$ where we expect $\FF_{\theta}(X)$ to be similar to $\FF_{\theta}(\TT(X, \alpha))$. Elements of the geometric transformation set $\alpha$ will guarantee semantic consistency with the input sample $X$. We show that many SOTA models will have $\FF_{\theta}(X) << \FF_{\theta}(\TT(X, \alpha))$ by a large margin (see Fig.~\ref{fig:fig2} and Fig.~\ref{fig:fig3}).

\begin{figure*}[htb]
\begin{minipage}{0.4\textwidth}
\begin{tikzpicture}
  \node (img)  {\includegraphics[scale=0.5]{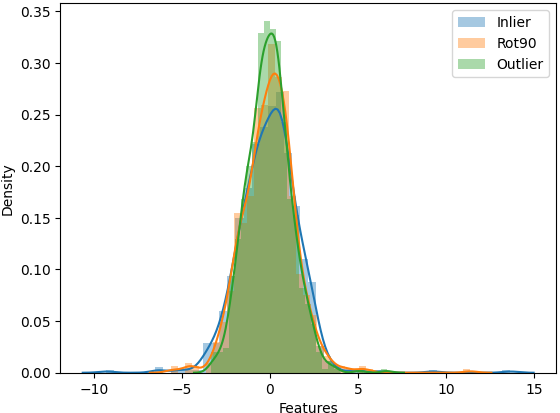}};
  \node[below=of img, node distance=0cm, yshift=1cm,font=\color{black}] {};
  \node[left=of img, node distance=0cm, rotate=90, anchor=center,yshift=-1cm,font=\color{black}] {};
 \end{tikzpicture}
\end{minipage}%
\begin{minipage}{0.4\textwidth}
\begin{tikzpicture}
  \node (img)  {\includegraphics[scale=0.5]{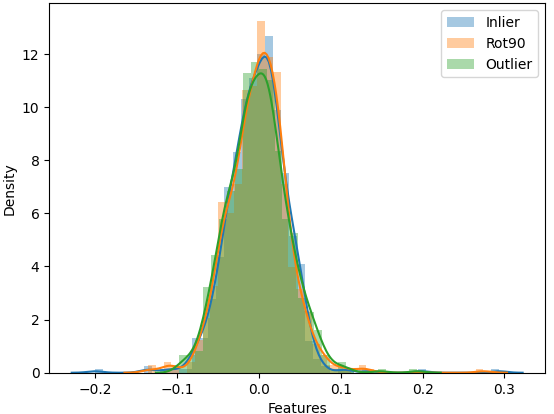}};
  \node[below=of img, node distance=0cm, yshift=1cm,font=\color{black}] {};
  \node[left=of img, node distance=0cm, rotate=90, anchor=center,yshift=0cm,font=\color{black}] {};
\end{tikzpicture}
\end{minipage}%
\caption{The figure on the left represents the feature vector distributions of an inlier, its 90-degree rotated version, and an outlier from ViT-B/16 before applying feature adaptation. As can be seen, the rotated inlier exhibits a significant change in feature vector values, resulting in an increased OOD score. In contrast, the figure on the right represents the same feature vectors after applying feature adaptation and demonstrates that both the inlier and its rotated version exhibit similar feature distributions. The images are drawn from the CIFAR-10 dataset.}
\label{fig:fig4}
\end{figure*}
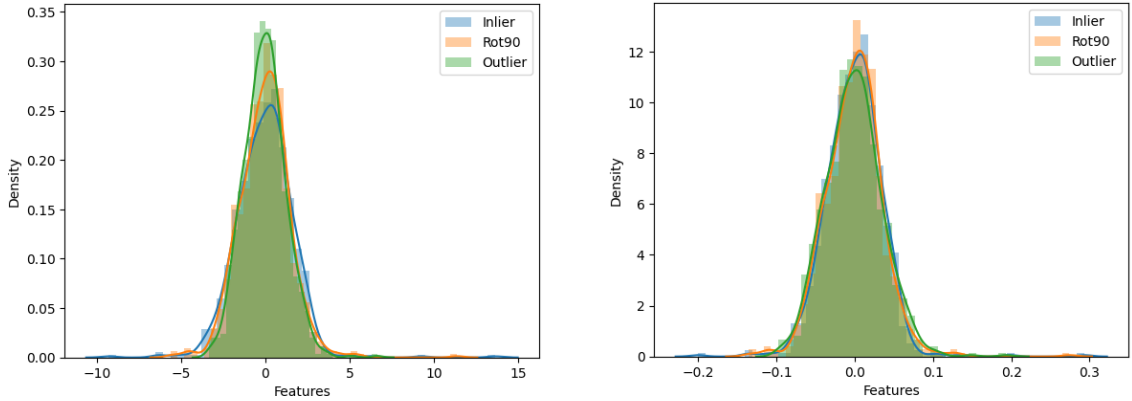

\begin{figure}
\begin{center}
  \includegraphics[width=0.42\textwidth]{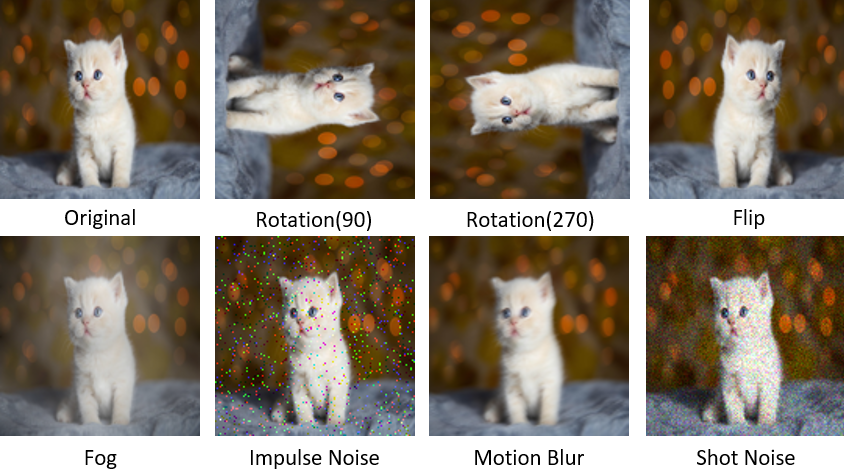}
\end{center}
\vspace{-4mm}
 
  \caption{To better represent real-world conditions, we propose a new OOD detection test setup. To create the datasets, we propose applying semantic-preserving transformations $\alpha$ to the inlier set $X$, as defined by $\TT(X,\alpha)$. The examples of these transformations are mentioned in section \ref{sec:method2}. Provided is the results of some of these transformations on an image of a cat. Note that the concept of a cat is preserved in all instances.
}
\label{fig:example}
\end{figure}

\textbf{Test Datasets: CIFAR-10-R, CIFAR-100-R and ImageNet-30-R}
\label{sec:method2}
We propose a new evaluation framework that better reflects the method's performance in the real world. This is achieved by creating test datasets that contain sufficient intra-class distribution shifts while maintaining semantic meaning. As such, we define the new test datasets as $Y' = \{(Y \cup B \cup \TT(B, \alpha))\}$ where $\alpha \in C$ or $\alpha \in A.$ $C$ is the set of common corruptions, and $A$ is the set of data augmentations. Following \cite{Hendrycks2019BenchmarkingNN}, elements in set $C$ include gaussian noise, shot noise, impulse noise, defocus blur, glass blur, motion blur, zoom blur, snow, frost, fog, brightness, contrast, elastic, pixelate, JPEG, speckle noise, gaussian blur, spatter and saturate, all of which are frequently encountered in natural images and maintain semantic consistency with the original image. Similar to \cite{Hendrycks2019BenchmarkingNN}, we also utilize five levels of severity when applying common corruptions to the image. Set $A$ consists of $90\degree$ rotation, $270\degree$ rotation, flip, random crop and resize (approximately 75\% to 80\% of the original image is preserved), and color jitter (brightness, contrast, saturation, and hue are set to 0.5), all of which intuitively maintain semantic similarity with the original image. Refer to Fig.~\ref{fig:example} to see examples of an original image and some of these transformations. Breaking down the resulting datasets, it consists of $Y$ and $B$ which is identical to the original test framework, and also $\TT(B, \alpha))$ which contains semantic-preserving transformations to create intra-class distribution shifts. If $\alpha \in \emptyset$ then the test dataset is identical to the existing framework.

\textbf{Generalizability Score (GS)}
\label{sec:method3}
With the definition of the new test datasets, we can investigate an evaluation metric that measures the ability of a model to generalize during OOD detection. Under the current OOD detection benchmark, we compute the Area Under the Curve (AUROC) for each class in a one-vs-rest setting, taking the average of AUROC across all classes as the measure of performance. However, in our proposed evaluation framework, we introduce a Generalizability Score ($GS$) defined as follows:

\begin{equation}
    \begin{split}
     \mathit{GS} = \FF_{\theta}(\TT(X, \alpha)) - \FF_{\theta}(X)
    \end{split}
 \end{equation}

GS is defined as the AUROC of a method tested on $Y'$ subtracted from the AUROC of the same method tested on $S$. $GS$ is designed to measure how well a method is able to generalize to intra-class distribution shifts. If the $GS$ is negative, then the method obtains lower AUROC on $S$ than $Y'$. If $GS$ is zero, then the method can be said to generalize perfectly to intra-class distribution shifts. If $GS$ is positive, then the method performs better on intra-class distributionally shifted data than the original testing framework. Therefore, a higher $GS$ indicates that the method being tested is more applicable to real-world applications.

\textbf{Adapting Deep Pre-trained Features for Our Framework:}
\label{sec:method4}
Upon examining the features extracted from deep pre-trained models to detect OOD data, it was revealed that specific alterations of samples, such as input rotations, can result in significant modifications to the output features. This sensitivity of these features to variations in inputs, which can lead to unpredictable changes in their outputs, undermines the reliability of the these models. To mitigate this problem, we propose to regularize these deep pre-trained features at test time to suppress unwanted changes. The core objective of this regularization is to curtail undesired feature modifications. By imposing a regularization penalty, we encourage the deep features to exhibit stability even under specific transformations, such as alterations in fine details. (See Fig. \ref{fig:fig4})

It is worth noting that while regularization techniques are conventionally integrated into the training process via loss function modifications, our proposal operates effectively even without retraining the model for the downstream task. This is achieved by applying regularization directly to the extracted features. A viable means of achieving this regularization is through the implementation of normalization techniques like L2. This normalization techniques scale the feature vector to have a specific norm or length. L2 normalization scales the feature vector such that the sum of the squares of its elements is equal to one. Mathematically, given a feature vector $v \in \mathbb{R}^n$, we can compute its L2 normalized version $v_{\text{L2}} \in \mathbb{R}^n$ as:

\begin{equation}
    v_{\text{L2}} = \frac{v}{\|v\|_2} = \frac{v}{\sqrt{\sum_{i=1}^n v_i^2}},
\end{equation}

Here, we divide each element of $v$ by the square root of the sum of the squares of its elements to obtain a normalized version of the feature vector that has an L2 norm of 1. Intuitively, this means that applying L2 normalization to the feature vectors makes the feature extractor more stable and well-behaved, and small changes in the input data lead to small changes in the feature vector.

In conclusion, this adaptation can help prevent the extracted features from becoming too large or too small, which can improve their stability and generalization performance. This technique proves especially advantageous when dealing with scenarios marked by limited data availability for the downstream task or when the pre-trained model has been trained on a distinct data distribution from that of the downstream task. By adopting the practice of normalizing extracted features, we markedly refine the performance and robustness of the models in the context of OOD detection. This process effectively mitigates the impact of feature scaling discrepancies or input data variations, transcending the perception of being a mere commonplace regularization technique.


\begin{table*}[htbp]
\centering
\caption{This table shows AUROC in \% for out-of-distribution (OOD) detection on CIFAR-10-R dataset. GS stands for Generalizability Score (the higher, the better), and the setups $\mathbf{U}$, $\mathbf{C}$, and $\mathbf{A}$  are Unrealistic, Corruptions and Augmentations, respectively. As shown, SOTA performance has saturated on the unrealistic setup, which is the existing OOD detection benchmark. But under transformation $\alpha$ which consists of common corruptions and augmentations, the performance drops drastically. After applying feature adaptation, we are able to recover the performance of deep pre-trained features under our new proposed evaluation framework.}
\label{table:CIFAR-10}
\resizebox{\textwidth}{!}{\begin{tabular}{|c|c|c|c|c|c|c|c|c|c|c|c|c|c|c|c|}
\hline
Setup & Models & Plane & Car & Bird & Cat & Deer & Dog & Frog & Horse & Ship & Truck & Mean & Mean(adapt.) & GS & GS(adapt.)\\

\hline

$\mathbf{U}$
& ResNet50 & 88.60 & 94.53 & 85.95 & 77.84 & 91.87 & 88.00 & 94.47 & 90.02 & 94.05 & 96.53 & 90.19 & 95.00 & N/A & N/A\\
& ResNet152 & 80.70 & 95.37 & 87.23 & 76.42 & 89.80 & 92.02 & 91.89 & 92.09 & 91.63 & 97.43 & 89.46 & 97.75 & N/A & N/A\\
& ViT-B/16 & 91.32 & 97.56 & 94.82 & 83.31 & 95.47 & 96.59 & 98.69 & 95.71 & 97.45 & 98.17 & 94.91 & 98.81 & N/A & N/A\\
& ViT-L/16 & 99.09 & 99.35 & 99.14 & 97.99 & 99.58 & 99.11 & 99.83 & 99.84 & 99.74 & 99.47 & \textbf{99.31} & \textbf{99.37} & N/A & N/A\\

\hline

$\mathbf{A}$
& ResNet50 & 81.70 & 85.99 & 80.33 & 76.82 & 83.35 & 81.27 & 93.22 & 75.26 & 84.50 & 88.73 & 83.62 & 88.37 & -6.57 & -6.63\\
& ResNet152 & 59.25 & 77.28 & 65.88 & 59.48 & 70.67 & 71.04 & 78.97 & 62.30 & 61.00 & 78.09 & 68.40 & 90.86 & -21.06 & -6.89\\
& ViT-B/16 & 76.08 & 81.52 & 80.82 & 73.17 & 82.41 & 83.95 & 94.36 & 70.05 & 76.61 & 82.91 & 80.19 & 92.98 & -14.72 & -5.83\\
& ViT-L/16 & 93.09 &  91.73 & 93.53 & 93.64 & 94.61 & 95.01 & 99.01 & 94.19 & 91.61 & 92.36 & \textbf{93.88} & \textbf{94.18} & \textbf{-5.43} & \textbf{-5.19}\\

\hline

$\mathbf{C}$
& ResNet50 & 81.09 & 83.42 & 77.55 & 73.52 & 87.27 & 78.34 & 91.60 & 79.94 & 85.70 & 86.37 & 82.48 & 84.06 & -7.71 & -10.94\\
& ResNet152 & 60.58 & 77.05 & 63.79 & 57.33 & 73.02 & 73.53 & 75.10 & 75.49 & 69.82 & 79.25 & 70.50 & 89.43 & -18.96 & -8.32\\
& ViT-B/16 & 78.06 & 88.15 & 81.15 & 69.66 & 87.35 & 86.50 & 91.44 & 86.68 & 87.95 & 89.44 & 84.64 & \textbf{94.99} & -10.27 & \textbf{-3.82}\\
& ViT-L/16 & 90.78 & 92.33 & 91.00 & 85.81 & 93.53 & 92.20 & 96.79 & 96.40 & 94.14 &  93.12 & \textbf{92.61} & 92.47 & \textbf{-6.70} & -6.90\\

\hline

\end{tabular}}
\end{table*}

\begin{table}[htbp]
\centering
\small 
\caption{This table reports the AUROC in \% for out-of-distribution (OOD) detection on CIFAR-100-R dataset (Super Class). We define three setups $\mathbf{U}$, $\mathbf{A}$, and $\mathbf{C}$ which represent the Unrealistic setup (existing OOD detection benchmarks), Augmentation setup, and Corruption setup. The same metrics and setups as table \ref{table:CIFAR-10} were used for this experiment.}

\label{table:CIFAR-100}
\begin{tabular}{|c|c|c|c|c|c|}
\hline
Setup & Models & Mean & Mean(adapt.) & GS & GS(adapt.)\\

\hline

$\mathbf{U}$
& ResNet50 & 89.18 & 94.33 & N/A & N/A\\
& ResNet152 & 87.60 & 96.78 & N/A & N/A\\
& ViT-B/16 & 93.26 & 97.83 & N/A & N/A\\
& ViT-L/16 & \textbf{98.34} & \textbf{98.51} & N/A & N/A\\

\hline
$\mathbf{A}$
& ResNet50 & 85.63 & 90.30 & \textbf{-3.55} & \textbf{-4.03}\\
& ResNet152 & 73.74 & 92.23 & -13.86 & -4.55\\
& ViT-B/16 & 83.56 & 93.39 & -9.70 & -4.44\\
& ViT-L/16 & \textbf{94.21} & \textbf{94.32} & -4.13 & -4.19\\

\hline
$\mathbf{C}$
& ResNet50 & 83.03 & 85.77 & \textbf{-6.15} & -8.56\\
& ResNet152 & 71.79 & 88.71 & -15.81 & -8.07\\
& ViT-B/16 & 83.26 & \textbf{93.54} & -10 & \textbf{-4.29}\\
& ViT-L/16 & \textbf{90.74} & 90.69 & -7.6 & -7.82\\

\hline

\end{tabular}
\end{table}

\begin{table}[htbp]
\centering
\small 
\caption{The presented table displays the \% AUROC values for detecting out-of-distribution (OOD) samples on the ImageNet-30-R dataset across three setups: Unrealistic (existing OOD detection benchmarks), Augmentation, and Corruption. The experiment utilized identical evaluation metrics and experimental setups as those outlined in previous tables.}
\label{table:imagenet}
\begin{tabular}{|c|c|c|c|c|c|}
\hline
Setup & Models & Mean & Mean(adapt.) & GS & GS(adapt.)\\

\hline

$\mathbf{U}$
& ResNet50 & 98.49 & 99.91 & N/A & N/A\\
& ResNet152 & 96.70 & 99.98 & N/A & N/A\\
& ViT-B/16 & 98.50 & \textbf{99.99} & N/A & N/A\\
& ViT-L/16 & \textbf{99.96} & 99.96 & N/A & N/A\\

\hline
$\mathbf{A}$
& ResNet50 & 94.45 & 99.51 & -4.04 & -0.40\\
& ResNet152 & 84.38 & 99.78 & -12.32 & -0.20\\
& ViT-B/16 & 94.97 & 99.78 & -3.53 & -0.21\\
& ViT-L/16 & \textbf{99.88} & \textbf{99.86} & \textbf{-0.08} & \textbf{-0.10}\\

\hline
$\mathbf{C}$
& ResNet50 & 95.13 & 99.46 & -3.36 & -0.45\\
& ResNet152 & 89.51 & 99.81 & -7.19 & -0.17\\
& ViT-B/16 & 96.26 & 99.82 & -2.24 & -0.17\\
& ViT-L/16 & \textbf{99.92} & \textbf{99.88} & \textbf{-0.04} & \textbf{-0.08}\\

\hline

\end{tabular}
\end{table}

\section{Experimental Results}
\label{sec:experiments}

\begin{table}
\centering
\caption{The table displays the AUROC values in percentage for OOD detection on the CIFAR-10-R dataset, focusing on one transformation (90-degree rotation) for the first class versus the rest on existing and our benchmarks. Our benchmark shows the performance drop of all SOTA models.}

 \begin{tabular}{|l|l|l||c|}
 \hline
SOTA Methods & Existing & Ours & Performance Drop \\
 \hline
DN2 \cite{Bergman2020DeepNN}                       & 83.06 & 66.28     & -16.78           \\
CSI \cite{Tack2020CSIND}             & 89.82                  & 52.12   & -37.70             \\
PANDA  \cite{mirzaei2022fake}            & 87.45                   & 75.99    &     -11.46   \\
FITYMI  \cite{Reiss2021PANDAAP}            & 99.23                   & 94.64    &     -4.59    \\
 \hline
\end{tabular}
\label{table:res}
\end{table}

The motivation behind leveraging deep pre-trained features is their richness in semantics due to their training on large and diverse datasets. In this section, we demonstrate how different architectures and models fail under our evaluation framework, highlighting that previous benchmarks are not a reliable indicator of a OOD detector's performance in real-world scenarios. Subsequently, we show how our proposed method can enhance their performance under our evaluation framework. Therefore, a paradigm shift in evaluation is necessary to guarantee the models' usability in practical settings.

In accordance with section \ref{sec:method2}, we utilize existing benchmark datasets in the creation of datasets in our evaluation framework. These include CIFAR-10 \cite{krizhevsky2009learning}, CIFAR-100 \cite{krizhevsky2009learning}, and ImageNet-30 \cite{hendrycks2019selfsupervised}. In the following, we provide descriptions and protocols defined on each dataset. {\bf CIFAR-10} has 32 x 32 RGB images of 10 natural objects and has been used for OOD detection. {\bf CIFAR-100} consists of 32 x 32 RGB images over 100 classes of natural objects. The 100 classes can be grouped into a 'coarse-grained' setting of 20 superclasses. We report performance using the coarse-grained setting. {\bf ImageNet-30} \cite{hendrycks2019selfsupervised} contains a subset of 30 classes of ImageNet \cite{deng2009imagenet} for one-vs-rest setting.

\textbf{Dataset Creation}
\label{sec:experiments2}
With all the previously mentioned datasets, we formulate three settings: Unrealistic ($\TT(B, \alpha), \alpha \in \emptyset$), Common Corruptions ($\TT(B, \alpha),\alpha \in C$), and Augmentations ($\TT(B, \alpha),\alpha \in A$). We name our created datasets by appending R to each source dataset name for OOD detection benchmarking. As seen in 
Tab.~\ref{table:dataset_stat}, our evaluation framework consists of the following datasets: CIFAR-10-R which contains the original 10000 test images, 50000 augmented images which were created by 5 different augmentations and 950000 images which were created with 19 different common corruptions with 5 levels of severity $(950000 = 10000 \times 19 \times 5)$. Creating CIFAR-100-R is also similar to CIFAR-10-R. ImageNet-30-R consists of the original test set of size 3000 plus 15000 augmented images (5 different augmentations) and 285000 images which were transformed with 19 common corruptions with 5 levels of severity ($285000 = 3000 \times 19 \times 5$).

\textbf{Testing Methodology}
\label{sec:experiments3}
Our testing protocol follows the one-class classification framework. During training, one class will be selected to be the set of inliers. When testing OOD detection using our framework, the same training class will be considered inliers in the test set, as well all semantic preserving augmentations and common corruptions will also be considered inliers. The rest of the test set which are the samples from other classes will be considered outliers. The results will be reported by averaging the AUROC of all classes.



\begin{table}
\centering
\caption{Our created dataset statistics. $\|A\|$ indicates the number of augmentation samples and $\|C\|$ shows the number of common corruption samples. The total represents the number of samples from the original test set and those created by transformations.}
 \begin{tabular}{|l|l|l|c|}
 \hline
Dataset & $\|A\|$ & $\|C\|$ & Total \\
 \hline
CIFAR-10-R                       & 50000 & 950000     & 1001000             \\
CIFAR-100-R               & 50000                  & 950000   & 1001000               \\
ImageNet-30-R               & 15000                   & 285000    &     303000      \\
 \hline
\end{tabular}
\label{table:dataset_stat}
\end{table}

\textbf{Evaluation and Discussion}
\label{sec:evaluation}
To showcase the deficiencies of current benchmarks in evaluating OOD detection models, we conducted an experiment using the CIFAR-10-R dataset. Given the computational challenges of using the entire dataset of over 1 million images for methods, we chose to focus on a proof-of-concept approach. Specifically, we selected a 90-degree rotation for the first class versus the remaining classes. Initially, we evaluated the performance of the SOTA OOD detection models on the existing protocol. Subsequently, we tested the models on test inlier images containing the rotated version of inliers. Although the models demonstrated favorable outcomes on the current benchmarks, our proposed benchmark uncovered a substantial performance drop across all SOTA models. This indicates the inadequacy of the current benchmarks in accurately determining OOD detection performance. (See Tab.~\ref{table:res})

Significant modifications to the output of deep models can occur when there are changes in the input caused by semantic-preserving transformations. These changes compromise the reliability and generalizability of deep models in real-world scenarios. To address this issue, we propose a novel approach that involves using feature adaptation to mitigate the effects of these transformations. Specifically, the proposed method utilizes deep pre-trained features that are post-processed to adapt for OOD detection to our new evaluation framework. The approach involves obtaining representations using multiple deep models with different architectures that are pre-trained on the ImageNet \cite{deng2009imagenet} dataset then they are adapted by regularizing features. Examples of these architectures include ViT-B, ViT-L, ResNet50 and ResNet152. Anomalies are identified based on the distance between each sample and its nearest normal training image, with larger distances indicating a higher likelihood of abnormality.

Tab.~\ref{table:CIFAR-10}, Tab.~\ref{table:CIFAR-100}, and Tab.~\ref{table:imagenet} show the results of SOTA deep models under existing benchmark datasets as well as our created datasets. Provided in the table are the class-wise AUROC scores, the average AUROC scores over all classes, and the Generalizability Score ($GS$) before and after applying our proposed adaptation method to indicate the method's applicability to real-world scenarios. Comparing the results on our datasets versus the existing benchmark datasets, we can clearly quantify the drop in performance of SOTA deep models under common corruptions and augmentations.

Fig.~\ref{fig:fig2} provides a more fine-grained study of the effects that some common corruption and augmentations have on the SOTA method's AUROC performance. The figure clearly shows the specific semantic-preserving transformations that cause distribution shifts such that SOTA methods fail, and by how much the performance will drop. The existence of such transformations is proof that SOTA methods fail to learn robust features from the inlier training data. Studying such distribution shifts is critical to choosing which real-world applications SOTA OOD detection methods can be applied to.

Similar to previous benchmarks, $\FF_{\theta}(X)$ typically has a value close to zero for an inlier sample $X_i$. Predictions closer to zero indicate an inlier prediction. In most cases, we observe that for an $X_i \in B$, $\FF_{\theta}(X_i)$ is significantly lower than $\FF_{\theta}(\TT(X_i, \alpha))$. In other words, the same OOD scoring function $\FF_{\theta}$ returns a significantly higher score on transformed samples. Samples with a score closer to one are more likely to be classified as OOD. Since both $X_i$ and $\TT(X_i, \alpha)$ are semantically in the inlier class, the gap between $\FF_{\theta}(X_i)$ and $\FF_{\theta}(\TT(X_i, \alpha))$ shows the error that the model will induce under transformation $\alpha$. As shown in Fig.~\ref{fig:fig3}, an ideal OOD classifier that is robust to intra-class distribution shift will have a scatter plot where all points follow the diagonal red line, where for an $X_i \in B$, $\FF_{\theta}(X_i) == \FF_{\theta}(\TT(X_i, \alpha))$.

By now, the disparity in performance between SOTA OOD detection methods in existing benchmarks and our proposed benchmark is clear. For this reason, we emphasize the AUROC metric with existing benchmarks is not representative of real-world performance. We believe future research must consider our framework for meaningful advancements in OOD detection research.

In the context of OOD detection, the ViT models performed well in both the Augmentation and Corruption setups, achieving the highest mean AUROC compared to the other models. This indicates that the ViT models are robust to different types of image perturbations and can accurately classify images even when they have been modified or are out-of-distribution. As shown in Tab.~\ref{table:CIFAR-10}, Tab.~\ref{table:CIFAR-100} and Tab.~\ref{table:imagenet}, The results showed that applying feature adaptation had a significant positive impact on models' performance, although the degree of improvement varied depending on the specific model and setup used.







\textbf{Can We Train on Proposed Transformations to Improve generalizability of OOD Detectors?}
\label{sec:question}
Including all possible transformations in the training step is not a practical solution to achieve high performance and generalizability on OOD detection tasks. In reality, set $\alpha$ can be of infinite size and it is impossible to completely cover all semantic-preserving distribution shifts during training. Even if we can form a subset of transformations in $\alpha$, the dataset creation and training will be computationally infeasible. In general, including such transformed samples during training does not solve the generalizability issue in OOD detection methods. Instead, there is a need to work on building more robust representations that can better generalize to semantic-preserving distribution shifts.

\section{Conclusion}
\label{sec:conclusion}
In this paper, we propose a new evaluation framework for OOD detection that is better suited for real-world scenarios and exposes the shortcomings of existing benchmarks. We introduce new OOD test datasets, including CIFAR-10-R, CIFAR-100-R, and ImageNet-30-R, along with a Generalizability Score (GS) that measures the performance differences between our proposed evaluation framework and existing benchmarks. Our results demonstrate that even state-of-the-art OOD detection models exhibit poor performance on our proposed benchmark, indicating that they are not suitable for deployment in real-world applications. To address this issue, we propose feature adaptation to enhance the performance of SOTA pre-trained models on our proposed benchmark and ensure the reliability of OOD detectors before deployment in real-world scenarios. It is clear that relying solely on AUROC on existing evaluation benchmarks is misguided because these datasets do not reflect the wide variety of distribution shifts that can occur in real-world scenarios. We suggest that future research should shift its focus towards evaluating the generalizability of OOD detection methods in real-world scenarios by utilizing our proposed evaluation framework. Reporting the GS can provide a more realistic indication of a method's performance and help researchers build more robust models.

{\small
\bibliographystyle{ieee_fullname}
\bibliography{egbib}
}

\end{document}